# A Hybrid Algorithm for Matching Arabic Names

Tarek El-Shishtawy
*Benha University, Faculty of Computers and Informatics*
t.shishtawy@ictp.edu.eg

**Abstract**

In this paper, a new hybrid algorithm which combines both of token-based and character-based approaches is presented. The basic Levenshtein approach has been extended to token-based distance metric. The distance metric is enhanced to set the proper granularity level behavior of the algorithm. It smoothly maps a threshold of misspellings differences at the character level, and the importance of token level errors in terms of token's position and frequency.

Using a large Arabic dataset, the experimental results show that the proposed algorithm overcomes successfully many types of errors such as: typographical errors, omission or insertion of middle name components, omission of non-significant popular name components, and different writing styles character variations. When compared the results with other classical algorithms, using the same dataset, the proposed algorithm was found to increase the minimum success level of best tested algorithms, while achieving higher upper limits .

## 1. Introduction

Information about individuals can be found in a variety of resources, such as population survey databases, national identifier databases, medical records, news articles, tax information, and educational databases. In all of these heterogeneous sources, name matching is a fundamental task for data integration that joins one or more tables to extract information referring to the same person.

Matching personal names have been used in a wide range of applications, such as record linkage or integration, database hardening, removing or cleaning up duplicated records, and searching the web. Unfortunately, name may not be known exactly, misspelled, or has spelling variations. Therefore, in these applications, the general word matching techniques were not enough, and optimized techniques have been developed to cope with matching multiple variations of the same personal name.

There are efficient and well established algorithms that deal with spelling errors variants for strings and name matching at a character level. For relatively short names that contain similar yet orthographically distinct tokens, character-based measures are preferable since they can estimate the difference between the strings with higher resolutions (Bilenko 2003). In languages where names has very close typographic structure, as Arabic, character level similarity is not enough to produce high precision matching.

Unfortunately, spelling errors are not the only source of name mismatching. People may report their names inconsistently by removing or inserting additional name tokens, adding initial titles, or writing different punctuation marks and whitespaces. In all of these cases, bag-of-words methods suit better to the matching problem, since they are more flexible at word level. On the other hand, token-based approaches are not able to capture the degree of similarity between individual tokens with minor variations in characters (Bilenko et al. 2003).

Experimental results show that hybrid techniques, which take word frequency as well as character based word similarities into account, increases matching A first attempt in this direction was introduced by Cohen et al. (2003), in the form of a measure called Soft-TFIDF, which extends Jaro-Winkler method to combine both of the frequency weight of words in a cosine similarity measure, and a CLOSE measure at the character level. Soft TFIDF algorithm works as follows: for each token Ai in the first name, find a word Bj in the second one that maximizes the similarity function. Moreau et al. (2008) showed that this may lead to double matching of words, and proposed a generic model to enhance the soft TFIDF. Camacho et. al. (2008) used a cost function that basically depends on matching all pairs of tokens and summing all edit distances at character level. The distance metric was modified by a frequency measure of the tokens. They also used a permutation factor (Monge-Elkan concept) to allow non-ordered sequence of word tokens to be matched.

In this work, we propose a hybrid sequential algorithm, that combines the advantages of token level and character level approaches to improve the name matching quality. The Hybrid algorithm is optimized for matching Arabic names. As we will discuss in section (3), Arabic names have a restricted writing order, close typographic pattern, subjected to middle token omission, and omission of common name tokens, even if occurred at the beginning of names. For these characteristics, existing algorithms cannot apply directly to matching Arabic names. Character-based hybrid algorithms may fail due the close typographic pattern, ignoring sequence order of tokens. Also, allowing permutation of tokens conflicts with restricted sequence of writing names. To improve the matching efficiency, most hybrid techniques include weights for token frequency, but don't give same attention to the relative position at which mismatch occurs.

The proposed hybrid algorithm is an extension to Levenshtein algorithm, with computing 'edit distances' at token level instead of character level in the basic algorithm. The sequential nature of the algorithm keeps the ordering importance of name tokens. The two names to be matched are considered as two bags of tokens, and the algorithm computes the cost of transform one bag into another. While the basic Levenshtein algorithm assigns a unity cost to all edit operations, the current algorithm assigns weights that reflect the importance of each edit operation. When matching pair of tokens, the importance of edit operation is determined by the relative position of the tokens in names, their frequency measure, and their character level partial similarity.

The remaining parts of this paper are organized as follows: In section 2, we present some basic techniques for name matching. After that, Section 3 briefly discusses the characteristics of Arabic naming system considered in our work. The proposed algorithm is detailed in section 4. The results of experimental comparisons are discussed in Section 5, Finally, conclusions are discussed in Section 6.

**2. Matching Algorithms for Names**

Name matching is the process of determining whether two name strings are instances of the same name. Multiple methods have been developed for matching names, which reflects the large number of errors or transformations that may occur in real-life data (Elmagarmid 2007).The basic goal of all techniques is to match names (or strings) that don't necessarily required to be identical. Instead of exact match, a normalized similarity measure is usually calculated to have a value between 1.0 (when the two names are identical) and 0.0 (when the two names are totally different). There are several well-known methods for estimating similarity between strings, which can be roughly separated into two groups: character-based techniques and token-based techniques. Levenshtein algorithm and its variants are character based matching techniques based on edit distance metrics. Levenshtein edit distance is defined originally for matching two strings of arbitrary lengths. It counts the minimum differences between strings in terms of the number of insertions, deletions or substitutions required to transform one string into the other. A score of zero represents a perfect match.

The basic Levenshtein method has been extended in many directions. [Hall and Dowling 1980]. For example, having an extension to consider reversals of order (transposition of characters) directly in the edit distance operation. Another direction of generalization is to allow different weights at character level. The weights for replacing characters can depend on keyboard layout or phonetic similarities (Snae 2007). In other research (Bilenko 2003), a distance function is produced by a distance function learner and the weights are learned from a training data set to have a combined record-level similarity metric for all fields.

The affine gap distance metric (Waterman 1976) offers a smaller cost for gap mismatch, and hence it is more suitable for matching names that are truncated or shortened. Smith and Waterman (1981) described an extension of edit distance and affine gap distance to find the optimal local alignment at character level. Regions of gaps and mismatches are assigned lower costs. Jaro (1989) [12,13] introduced a string comparison metric, which is dependent on both number of common characters and number of non-matching transposition in the two strings.

Token based approaches are motivated by the fact that most of the differences between similar named entities often arise because of abbreviations or whole-word insertions and deletions, and hence this model should produce a more sensitive similarity estimate than character based approaches. Also, experimental results show that including token frequency as a parameter in matching algorithms lead to a significant improvement in matching accuracy. Jaccard and vector space cosine similarity are examples of difference measures that operate on tokens, treating a string as a "bag of words". In these approaches, the two string names to be compared are divided into sets of words (or tokens), then a similarity metric is considered over these sets.

The Jaccard similarity between the word sets A and B is defined as the size of the intersection divided by the size of the union of the sample sets: $J(A,B) = |A \cap B| / |A \cup B|$. The algorithm has been extended to compare bi-grams (paired characters of two string), tri-grams or n-grams. Strings can be "padded" (Keskustalo 2003) by adding special characters at the beginning and end of strings, Padded n-grams will result in a larger similarity measure for strings that have the same beginning and end but errors in the middle.

TFIDF or cosine similarity is another measure, widely used in the information retrieval community. The basic TFIDF makes uses of the frequency of terms in the entire collections of documents and the inverse frequency of a specific term in a document. The TFIDF weighting method is often used in the Vector Space Model together with Cosine Similarity to determine the similarity between two documents. Similarity between database strings, or between a database string and a query string, is then computed using the cosine similarity (inner product) of the corresponding weight vectors, essentially taking the weights of the common tokens into account. TFIDF similarity of two word sets A and B can be defined as

$$\text{TFIDF}(A,B) = \sum_{w \in A \cap B} V(w,A).V(w,B) \qquad (1)$$

V is a weight vector that measures normalized TFIDF of word w. Like Jaccard, the TFIDF scheme depends on common terms, but terms are weighted; these weights are larger for words w that are rare in the collection of strings from which A and B were drawn. The basic TFIDF does not account for misspelling mistakes in words. Cohen et al.(2003) proposed a soft TFIDF with a heuristic that accounts for certain kinds of typographical error.

Soft TFIDF is one approach that combines both string-based and token-based distances. In this approach similarity is affected not only by the tokens w that appear in sets A and B, but also for those "similar" tokens in A that appear in B.

$$\text{soft TFIDF}(A,B) = \sum_{w \in \text{close}(\theta, A, B)} V(w,A).V(w,B)\, D(w,B) \qquad (2)$$

Where D is the character based distance of the word w, such that it is greater than a threshold θ. The new set close allows to integrate a token based distance and the statistics of a particular corpus in the similarity evaluation of a particular word.
Both of TFIDF and Soft TFIDF are insensitive to the location of words, thus allowing natural word moves and swaps (e.g., "John Smith" is equivalent to "Smith, John"). However this is useful in many naming systems, it doesn't fit Arabic naming system which is characterized by restricted component order. Camacho et al. (2008) proposed a similar metric that combines both the frequency of words, and the edit-based distances of each word pairs of the two names. Also, strings may be phonetically similar even if they are not similar in a character or token level. Soundex (Holmes and McCabe 2002), Phonex, and Phonix (Gadd 1990) are examples of phonetic based techniques that convert the name into a sequence of codes that represent how the name is spoken. Phonetic representation of the names are used either for exact (or approximate) match.

When considering context information stored with names (such as address, mail, and other details) to increase the likelihood of a match, the problem is called data or record linkage (Xiao 2011). Many techniques were proposed for record linkage, where not only pairs of name strings are matched, but many other matching features (Monge and Elkan 1996). Winkler (2002) demonstrated how machine-learning methods could be applied in record linkage situations where training data were available. Name is time-independent information, and therefore, even in feature-based approaches, having an effective name matching is crucial (Winkler 2006). This work considers only name matching without taking any context information into account.

## 3. Characteristics of Arabic Name Variations

Exact string matching of personal names is problematic for all languages because names are often queried in a different way than they were entered. The proposed algorithm deals with the following problems concerning Arabic names..

### 3.1 Very Close Typographic Structure

Spelling errors normally occur during data entry. It may be due to typographical errors, cognitive errors, or phonetic errors. Whatever is the reason of the error, the source and target names are considered strings differed at a character level. According to Jurafsky and Martin (2003), this type of error can be categorized as insertion, deletion or omission, substitution, and transposition. There are efficient and well established algorithms that deal with spelling errors variants for string. When data is represented by relatively short strings that contain similar yet orthographically distinct tokens, character-based measures are preferable since they can estimate the difference between the strings with higher resolutions.
The reason that misspelling errors are particularly difficult in Arabic names is the close typographical structure of names. For example, inserting the character "و" to the name "محمد", yields a correct name "محمود". Also, substituting the character "أ" with "م" in "محمد", gives a correct name "أحمد". If Levenshtein algorithm is used for matching two names with a length of 20 characters of each name, a single edit distance will show 95% matching similarity of the two names, while they are two different persons. The problem is how to know if the name is written incorrectly or refer to another person (eg., his brother), especially when searching family databases.

### 3.2 Omission of name components

While it is common to have one first, one or more middle, and a surname name for writing person name, several variations exist in real free form names. The same problem exists in many other languages and it is reported by Borgman and Siegfried (1999) that, there are no legal regulations of what constitutes a name. The source of the ambiguity, in many cases, are people themselves as they report their names differently depending upon the organization they are in contact. Examining different Arabic databases show that name omission is a serious problem that should be handled efficiently in any Arabic name matching algorithm. Name omission is related to both of position and frequency as follows:

*3.2.1 Name Order*

*Persons tend to write their names in a restricted correct order. They may omit one or more tokens, but still keeping correct order.* Examining different writing styles of Arabic names, shows that transposition errors occurred rarely. Therefore, one wants "Hamed Mohamed Fawzy Ibrahim" to match with "Hamed Mohamed Ibrahim" but not with "Hamed Fawzy Mohamed Ibrahim". The built in sequential nature of the proposed algorithm assigns one edit distance to 'omission' and 'two edit distances' to transposition.

*3.2.2 Position of omitted name*

*Persons tend to omit one or more middle names, while fewer name omissions typically occur at the beginning or at the end of names.*
The analysis show that a person is keen on writing his first and surname carefully. This raises the position importance of the name variations. For example, one wants "Hamed Mohamed Fawzy Ibrahim" to match with "Hamed Mohamed Ibrahim" but not with "Mohamed Fawzy Ibrahim". To realize position relation, the proposed algorithm gives less importance to name omission – or insertion - that happened at the middle of the name, and more importance to first and last names.

Table 1 Arabic Common Name Frequency

| **Arabic Name** | **English name** | **TF** |
|---|---|---|
| محمد | **Mohamed** | **11.38%** |
| احمد | **Ahmed** | **5.98%** |
| محمود | **Mahmoud** | **2.39%** |
| على | **Ali** | **2.28%** |
| ابراهيم | **Ibrahim** | **2.07%** |
| حسن | **Hassan** | **1.84%** |
| السيد | **Alsayed** | **1.54%** |
| مصطفى | **Mostafa** | **1.33%** |
| حسين | **Hossien** | **0.87%** |
| **Total percentage of top common Arabic name tokens** | | **29.7%** |

*3.2.3 Frequency of omitted name*

*Persons tend to omit non-significant components of their names, i.e., omission is likely to occur with common names.*
Results of analyzing a sample of 8140 Egyptian full names show that nearly 30% of all name components lies within a set of only 9 common names as shown in table (1). In the proposed algorithm less importance is given to a common name omission or insertion. For example, one wants "Hamed Mohamed Fawzy Ibrahim" to prefer the match with "Hamed Fawzy Ibrahim" than "Hamed Mohamed Ibrahim", because 'Mohamed' is not an indicative name as ' Fawzy'.

Common names have another impact when searching a web for famous persons. When searching for the former president of Egypt, many people don't know that his first name is 'Mohamed', and search the web only for 'Hossny Mubark". The search engine should be smart enough to return also matches with his full name as top hits, because 'Mohamed' is a common name and it is expected to be omitted. This is different from returning 'Gamal Hossny Mubark' – his son - since 'Gamal' is not a common name.

The previous discussion shows that the frequency distributions of name values can be used to improve the quality of name matching. They can either be calculated from the data set containing the names to be matched, or from a more complete population based database like a telephone directory or an electoral roll.

**3.3 Writing Styles Character Variations**

One important component of the proposed work is name standardization. Standardization eliminates writing styles character variations and hence, makes the data comparable and more usable.  To produce a uniform representation, the algorithm runs SQL script to replace various spelling of words with a single spelling. For instance, different prefixes, spacing, punctuations, and character variations are replaced with a single standardized spelling

Name standardization (or character variation elimination) module is common module in name matching algorithms (Patman and Thompson 2003). In the current work, standardization concept is optimized for Arabic names. For instance, the module trims certain prefixes such as ( دكتور م. أ.د. د, /السيد ، /د), etc ), replaces multiple blanks with a single blank replaces the characters (أ، إ، آ) with ا), replaces the ending character (ي) with (ى). There are some cases where Arabic name component is composed of two tokens. For example, a prefix name component (عبد) and a postfix name component (الدين) can't be standalone names. There is no standard style for writing composite names, as it is not always to have a distance space between them. To standardize composite names, either a leading or trailing spaces are removed, whenever a pre/post tokens are detected.  Name style standardization is an inexpensive step, but improves the overall performance for name matching.

**4. The proposed algorithm**

We started with the Levenshtein edit distance similarity metric and extended it to handle name matching at a token level. The Sequential nature of Levenshtein method ensures that the sequential name order of tokens is considered. Following the variant of Needleman-Wunch (gap cost), the current algorithm replaces the fixed unity cost of the simple Levenshtein form, with a cost function that is dependent on frequency and position of tokens to be matched.

Specifically, the implementation of the proposed algorithm applies three modifications to the basic Levenshtein distance metric. The first modification is the application of the same dynamic programming technique at token level instead of character level in basic Levenshtein. For example the distance between the two names a = ('Mohamed', 'Ahmed', 'Hassan', 'Ali') and b=('Mohamed', 'Hassan', 'Ali', 'Ibrahim') is two. This is because (a) requires two edit operations (deletion of token 'Ahmed', and insertion of token 'Ibrahim' at the end of a).

The second modification is the mapping of the frequency and position importance of name tokens – discussed in sections 3-2 and 3-3 - with a cost function C, instead of assigning fixed unity cost for all edit operations. The role of the cost function C is to lighten (or strengthen) the effect of token mismatch according to word position and frequency.

The third modification is the implementation of partial matching of individual token pairs at the character level. This fine grained level ensures that pairs which have slightly different misspellings are not ignored.

For a two tokens $(a_k, b_l)$, where $1 \leq l \leq L$ and $1 \leq k \leq K$

$$H(k, l) = min \begin{cases} H[k-1, l] + C_{k,l} \\ H[k, l-1] + C_{k,l} \\ H[k-1, l-1] + \text{TokenCost}(a_k, b_l) \, C_{k,l} \end{cases} \quad (3)$$

Where $C_{k,l}$ is given by

$$C_{k,l} = P_{k,l} \, F_k \quad (4)$$

$P_{k,l} \, F_k$ are the position and frequency costs

TokenCost is the token-pair similarity cost at a character level. The final similarity percentage between the two name strings is then given by

$$\text{sim}[a, b] = 1 - \frac{H[K, L]}{\max(K, L)} \quad (5)$$

**4.1 Token Pair Mismatch Cost**

The 'TokenCost'-measure is the cost of partial match between tokens a[k] and b[l], which captures word spelling errors at character level. Pairs of tokens that are not necessarily identical are also considered in the edit operation but with a cost that depends on their similarity. The proposed 'TokenCost'-measure combines the token level edit operations with approximate token matches. The concept is similar to the 'close' function in soft TFIDF. In our implementation, 'TokenCost' function has a value that ranges from zero (for exact token match, and hence has 0 required edit actions), to 1 (for completely mismatch, and hence has 1 complete edit action). When similarity is below a certain threshold, the pairs are considered dissimilar and distance function is set to one. It is computed with Levenshtein edit distance, and clipped at a threshold value θ. The threshold limit is implemented with the threshold rule:

$$\text{TokenCost}(a_k, b_l) = \text{Levenshtein}(a_k, b_l)$$

IF Levenshtein $(a_k, b_l) \geq \theta$ then TokenCost$(a_k, b_l) = 1$ (6)

For two names a and b, the current algorithm defines a Levenshtein cost of the two token words $a_k \in a$ and $b_l \in b$ as $0 \leq$ Levenshtein $(a_k, b_l) \leq 1$. Distance cost ranges from zero for complete match to one for complete mismatch. The distance depends on Levenshtein edit metric. The basic Levenshtein algorithm is a character based approach, which takes two strings, $a_k$ and $b_l$ of lengths m and n characters, and returns the Levenshtein distance between them. For all i and j, d[i,j] will hold the Levenshtein distance between the first i characters of $a_k$ and the first j characters of $b_l$. The elements of d[i,j] is computed according to:

If $a_k[i] = b_l[j]$ then $d[i, j] = d[i-1, j-1]$

$$\text{Else } d[i, j] = min \begin{cases} d[i-1, j] + 1 & \text{// a deletion} \\ d[i, j-1] + 1 & \text{// an insertion} \\ d[i-1, j-1] + 1 & \text{// a Substitution} \end{cases}$$

(7)

The similarity measure is then given by

$$\text{Levenshtein}[a_k, b_l] = \frac{d[m,n]}{\max(M,N)} \qquad (8)$$

**4.2 Position mismatch cost of tokens**

As Western naming systems, Arabic name token's order is very important. Family name usually appears as the last token of the name. Therefore, any successful name matching algorithm should allow for gaps of unmatched characters (eg., Smith-Waterman algorithm) and the problems of out of order of tokens (eg. Monge-Elkan method). The proposed algorithm satisfies both constraints with more flexibility for adjusting the relative importance of token position.

Instead of using the number of edit operations as a distance metric, the proposed algorithm uses the cost of the edit operations required to transform one string to another. When matching complete names, inserting or deleting a token (name) at the beginning (first name) or at the end (family name) of a complete name have different costs of doing same edit actions for middle names. We call this position cost, which is implemented using a position weight cost $0 \leq P \leq 1$.

$P_{k,l}$ is the position weight for matching token k with token j. In general, the proposed position cost is flexible and can have different values at each token position. In our implementation, persons are keen to write their first and last names. and for this reason, more importance is given when matching first and last names. Position weight is assigned a complete unity edit cost when matching either the two first or the two last tokens in both strings.

In our implementation, the position cost rule is given by

$$P_{k,l} = \begin{cases} 1, & (k=1 \text{ and } l=1) \\ & \text{or } (k=K \text{ and } l=L) \\ \beta & \text{Otherwise} \end{cases} \qquad (9)$$

Position rule is used to initialize the zero rows and columns as shown in Table 1. Note that this initialization is different from Smith and Waterman [34], which assigns zeroes to all zero rows and columns.

Table (2) illustrates an example of computing H(l,k) when considering only position weights to match two strings a [L]= (W1, W2, W3, W4, W5) and name b[K]= (W1, W3, W4, W6). The total distance cost is (1+ β) /5, due to cost of a single token omission (β) and the cost of last word mismatch (1). By varying the value of β we can control the position cost of the token.

Table 2 Common Name Frequency

|    |          | W1  | W3  | W4   | W6    |
|----|----------|-----|-----|------|-------|
|    | 0        | 1   | 1+β | 1+2β | 2+2β  |
| W1 | 1        | 0   | β   | 2β   | 3β    |
| W2 | 1+β      | β   | β   | 2β   | 3β    |
| W3 | 1+2β     | 2β  | β   | 2β   | 3β    |
| W4 | 1+3β     | 3β  | 2β  | β    | 2β    |
| W5 | 2+3β     | 4β  | 3β  | 2β   | 1+β   |

**4.3 Frequent Name mismatch cost**

Motivated by the assumption that 'persons tend to omit their common names', we allow the cost of edit distances for a common names to have different cost than rare names. This called a term frequency weight $0 \leq F \leq 1$, and is given by

$$F_l = 1 - \alpha \, \frac{TF(a_l)}{MTF} \qquad (10)$$

Where $TF(a_l)$ is the Term Frequency of token $a_l$ MTF is the maximum term frequency of names, and α is a frequency weighting factor. To allow the algorithm to have a maximum frequency effect, α is set to 1. In this case, the common name frequency cost, and hence the overall cost of the edit operation, is nearly 0.

**5. Experimental Results**

**5.1 Datasets**

We have used two types of datasets: the base set and test sets. The base dataset contains a sample of 8140 names extracted from MIS of Egyptian Universities staff members. To make the base dataset usable, all names are standardized to eliminate writing styles character variations as explained in section 3.3. The extracted sample contains two fields: 1) a Base name IDentifier B_ID, and the Base Name (BName). The experiment uses six different test datasets, with each containing 300 names extracted randomly from the base dataset. Each test set is subjected to a noise to induce one of the following errors: 1) deletion of random single character, 2) Deletion of random two characters, 3) omitting first token , 4) omitting second token, 5) omitting third token, or 6) omitting both second and third tokens. Each test set has one type of distortion, but all test sets have similar field structure: 1) Test name Identifier (T_ID), 2) the distorted name (DName), and 3) a Reference to the original base name Ref_B_ID. Then we have six test sets for which true match status is known with each carrying one type of errors.

**5.2 Experimental Methodology**

The purpose of all experiments is to measure the degree by which the proposed algorithm can overcome each type of inserted noise. The 300 distorted names of each test set are matched against the original 8140 base set. Since we interested to automate the matching process, we define 'success' as a direct measure of the algorithm performance. The testing environment was adopted to accept only the top-scored name from the tested algorithm as a match. The match is then counted as true, if it corresponds to its original name in base set. The success percentage is calculated as the count of true-matches divided by test set size.

For a single distorted name (DNname, Ref_B_ID), the test methodology consults the tested algorithm (the proposed algorithm or other algorithms), to compute the similarity against the base set. The test environment keeps the running maximum similarity and the Base name ID (Sim, B_ID) as a candidate matched Name. This match is true when the final maximum similarity name is the same as the original base name (i.e., B_ID = Ref_B_ID). For a given error type, the process is repeated for the 300 distorted names The algorithm is shown in figure 1.

```
Algorithm Success-Match-Percentage
Input :    Specific values of θ, α and β
           Specific Distorted Data Set DDSᵢ (DNname, Ref_B_ID) ,  1 ≤ i ≤ 300
           Base Data Set BDSⱼ (BName, B_ID), ,  1 ≤ j ≤ 7140
Begin
  True Match Count = 0;
  For (I = 1 to 300) {
         Max Running score = 0;
         Ref record = 0;
         For (j=1 to 7140) {
                Apply equation (5) to get Sim (DName(i), BName(j)
                  If Sim> Running max Score {
```

```
                Set Max Running Score = Sim;
                Set Ref Record = B_ID (j); }
        } End loop j
        If  Ref record = Ref_B_ID (i) {   /* success */
            Increment True Match Count by 1 ; }

  } End loop i
  Success percentage = True Match Count / 300;
End;
```

Figure 1: The methodology for testing the proposed and other algorithms.

In summary, we have used artificially generated files with each carrying one type of error. Although the generated data sets do not approximate the types of errors that occur in real data, they are very useful in: 1) analyzing the effect of different parameters on each type of error, 2) determining the upper and lower success limits of each algorithm.

**5.3 Results**

We ran our experiments on data sets, with each carrying only one type of error. Results of the proposed algorithm are summarized in Tables 3 to 8. Each cell entry is the percentage of success which is calculated as the number of true matches of the test set names divided by the total number of names in the set, To illustrate the role of position weight and individual token match threshold in evaluating a name matching, the experiment was repeated for different values of β and θ.

In general, the results show that the algorithm succeeds to overcome single character and second token omissions with nearly 100% when properly setting the parameter values. Names with two characters omission is matched successfully with 97.33%. Following is the third token omission with success percentage of 95.00%. Table () shows that the first token omission is the serious error, for which the algorithm succeeds only with 89% to match correct names. A worst accuracy of 79.67% is achieved when a person omits both of his/her both second and third names.

*5.3.1 Effect of threshold and position weights*

The results show that there is a conflict need for the value of θ required for obtaining optimum results for all types of errors. Character omission errors require a moderate value (θ = 0.5), while overcoming token omission requires lower values of θ ranging from 0 to 0.1. It is clear that for lower when thresholds, the similarity of token cost at character level (third term of the minimum relation of equation 3), always dominates, and the system behavior is mainly a character level operation. On the other hand, setting threshold θ = 1, a unity edit cost is assumed, and the algorithm neglects all character variations at token level.
The position weight β required to obtain best results have a fixed pattern for both characters and token omission errors. Better results are usually found when β ranges from 0.5 to 0.7. The only exception is the first token omission error which needs higher values of β. To explain the effect of β values, let's return to equation (3). If β has a maximum value of 1, no positional information will be used since a unity edit cost is assumed whatever is the position, and the algorithm behavior will give equal importance to all token variations, which correspond to a static string distance. Also, for very low values of β, the algorithm nearly neglects all token level variations but first and last tokens.

Table (3): One character omission success percentages

| θ | β | | | | | |
|---|---|-----|-----|-----|----|---|
|   | 0 | 0.1 | 0.3 | 0.5 | 07 | 1 |

| 0   | 50.33% | 70.33% | 91.00% | 93.67% | 94.67% | 94.00% |
|-----|--------|--------|--------|--------|--------|--------|
| 0.1 | 50.33% | 70.33% | 91.00% | 93.67% | 95.67% | 94.00% |
| 0.3 | 82.33% | 91.00% | 95.33% | 95.33% | 97.67% | 95.33% |
| 0.5 | 87.67% | 97.00% | 97.33% | 99.00% | 99.00% | 99.33% |
| 0.7 | 88.00% | 97.33% | 98.67% | 98.67% | 98.00% | 98.33% |
| 1.0 | 87.33% | 97.00% | 98.67% | 98.67% | 98.00% | 98.00% |

Table (4): Two characters omission success percentages

| $\theta$ | $\beta$ | | | | | |
|---|---|---|---|---|---|---|
|     | 0      | 0.1    | 0.3    | 0.5    | 07     | 1      |
| 0   | 29.33% | 39.00% | 64.00% | 70.67% | 71.33% | 72.00% |
| 0.1 | 29.33% | 39.00% | 64.00% | 79.67% | 90.67% | 82.00% |
| 0.3 | 67.33% | 78.33% | 88.67% | 90.33% | 94.67% | 91.00% |
| 0.5 | 79.67% | 92.00% | 96.00% | 96.33% | 97.33% | 96.67% |
| 0.7 | 78.67% | 94.33% | 96.67% | 96.67% | 96.33% | 95.33% |
| 1.0 | 78.67% | 94.33% | 96.67% | 96.67% | 95.67% | 95.00% |

Table (5): First token omission success percentages

| $\theta$ | $\beta$ | | | | | |
|---|---|---|---|---|---|---|
|     | 0     | 0.1    | 0.3    | 0.5    | 07     | 1      |
| 0   | 2.33% | 34.00% | 74.33% | 82.33% | 87.00% | 89.00% |
| 0.1 | 2.33% | 34.00% | 74.33% | 82.33% | 87.00% | 89.00% |
| 0.3 | 2.67% | 28.67% | 66.33% | 76.67% | 85.33% | 88.67% |
| 0.5 | 4.00% | 16.33% | 53.00% | 67.33% | 81.33% | 84.67% |
| 0.7 | 4.00% | 10.00% | 41.33% | 63.67% | 78.67% | 82.00% |
| 1.0 | 4.33% | 8.33%  | 37.00% | 62.33% | 76.67% | 79.33% |

Table (6): Second token omission success percentages

| $\theta$ | $\beta$ | | | | | |
|---|---|---|---|---|---|---|
|     | 0      | 0.1    | 0.3    | 0.5    | 07     | 1      |
| 0   | 92.33% | 99.33% | 99.33% | 99.67% | 99.67% | 99.00% |
| 0.1 | 92.33% | 99.33% | 99.33% | 99.67% | 99.67% | 99.00% |
| 0.3 | 92.33% | 99.00% | 98.67% | 98.00% | 97.33% | 96.67% |
| 0.5 | 92.33% | 98.67% | 98.33% | 97.33% | 96.00% | 93.33% |
| 0.7 | 92.33% | 98.00% | 97.67% | 97.00% | 95.67% | 90.00% |
| 1.0 | 92.33% | 97.33% | 97.00% | 96.67% | 95.33% | 87.67% |

Table (7): Third token omission

| θ | β | | | | | |
|---|---|---|---|---|---|---|
| | 0 | 0.1 | 0.3 | 0.5 | 07 | 1 |
| 0 | 77.00% | 83.67% | 91.00% | 94.33% | 95.00% | 71.00% |
| 0.1 | 77.00% | 83.67% | 91.00% | 94.33% | 95.00% | 71.00% |
| 0.3 | 77.00% | 83.33% | 89.00% | 90.67% | 93.67% | 65.67% |
| 0.5 | 77.00% | 83.33% | 85.67% | 89.33% | 93.00% | 60.67% |
| 0.7 | 77.00% | 83.33% | 84.67% | 89.00% | 92.67% | 55.00% |
| 1.0 | 77.00% | 82.33% | 83.00% | 88.00% | 92.00% | 51.67% |

Table (8): Second and third tokens omission success percentages

| θ | β | | | | | |
|---|---|---|---|---|---|---|
| | 0 | 0.1 | 0.3 | 0.5 | 07 | 1 |
| 0 | 76.67% | 76.33% | 79.67% | 79.67% | 80.33% | 71.00% |
| 0.1 | 76.67% | 76.33% | 79.67% | 79.67% | 80.33% | 71.00% |
| 0.3 | 76.67% | 76.00% | 78.00% | 78.00% | 77.00% | 65.67% |
| 0.5 | 76.67% | 76.00% | 78.00% | 78.00% | 74.00% | 60.67% |
| 0.7 | 76.67% | 76.00% | 78.00% | 78.00% | 73.67% | 55.00% |
| 1.0 | 76.67% | 75.67% | 77.33% | 77.33% | 72.00% | 51.67% |

*5.3.2 Comparing results with other algorithms*

We now present an experiment comparing our hybrid weighting technique (run at α =1, β=0.7, θ =0.1) with different state-of-the-art systems using the same Arabic data sets and the same experimental methodology. The tested algorithms are: Levenshtein, Monge-Elkan, Jaro-Winkler (JW), and Soft-TFIDF (run at JW, θ =0.9). The Java open-source toolkit of tested algorithms is available at http://secondstring.sourceforge.net/) (Cohen et al. 2003).

When considering only character misspelling errors, the basic Levenshtein algorithm does the best, followed by Jaro-Winkler with a success of 95%. Both of the proposed algorithm and Soft TFIDF have the same average success percentage of 93% for character level errors, while Monge-Elkan algorithm does the worst accuracy result, where nearly 45% of the distorted names are matched incorrectly.

For token omission errors, as shown in Table 9, our proposed algorithm seems generally the best when considering either the success boundary or individual token's errors. It has a better success range (from 80.3% to 99.7%) than Soft TFIDF (from 69.0% to 94.3%). A part of this better performance may be due to missing prior Arabic frequency knowledge for Soft TFIDF algorithm. The next best performance for the tested algorithms is Monge-Elkan method, which has a success boundary ranging from 23.0% to 89.3%. The worst result comes from Jaro-Winkler with success ranging from 8.3% to 72.3%.

It is important to note that the actual success percentage of all presented algorithms will be at a point between their upper and lower boundary limits in real data sets. The actual success operation of any algorithm depends also on the mixture amount of different types of errors in the real sets. Highly spanned algorithm

Table (9): Comparison of success percentages of the proposed algorithm with existing matching algorithms using Arabic dataset.

| Omission Error Type | Proposed Average β=0.7 θ =0.1 | Basic Lev. | Monge Elkan | Jaro Winkler (JW) | Soft TFIDF JW θ =0.9 |
|---|---|---|---|---|---|
| One character | 95.7% | 100% | 66.7% | 96.0% | 95.7% |
| Two characters | 90.7% | 100% | 44.3% | 94.0% | 90.3% |
| First token | 87.0% | 90.7% | 89.3% | 72.3% | 86.0% |
| Second token | 99.7% | 67.7% | 40.3% | 68.7% | 94.3% |
| Third token | 95.0% | 65.3% | 35.0% | 51.3% | 91.3% |
| Second & third | 80.3% | 16.0% | 23.0% | 8.3% | 69.0% |

**7. Conclusions**

In this work, the basic character based Levenshtein approach has been extended to token-based distance metric. The algorithm is enhanced to combine the minor misspellings differences at the character level, and both of position and frequency parameters at token level.

The experimental results demonstrate that taking position weight β, character threshold θ into account to weight distance metrics improves the performance of name matching. The proper value for θ is very critical since it determines the granularity level behavior of the algorithm. For example, If θ is set to one, the algorithm ignores all individual token's spelling errors, and tends to be a token level algorithm. On the other hand, setting θ to zero, the algorithm considers all token pairs to be matched with a similarity score, and hence the operation of the algorithm approaches character level algorithms. Sets which are characterized by higher spelling errors need higher thresholds, while those that are characterized with token omissions need lower thresholds. However, our presented hybrid name matching algorithm is able to work efficiently under different types of errors. Due to flexibility in changing the algorithm parameters, it can be optimized according to dataset characteristics to obtain better results. Comparing the performance of different name matching algorithms is still a problematic issue. It was agreed that even metrics that demonstrate high performance for some data sets can perform poorly on others [2]. In real environment, one cannot expect the matched set characterization. Therefore, in this work, we follow another comparison methodology by setting upper and lower limits of success of tested algorithms based on variety of errors schemes they may be faced in real sets. Clearly highly spanned success algorithms are better and will have robust performance for different datasets. Also, since we are intended to automate the whole matching process, we defined a 'success' of tested algorithm as its ability to produce the correct matched name as a top scored name. Using this methodology with a large Arabic dataset, the average performance of the proposed algorithm was compared with other classical algorithms. It was found that, it raises the minimum success level from 69% - of best tested Soft TFIDF algorithm - to about 80%, (for second and third token omission), while achieving an upper accuracy limit of 99.67%. The best accuracy is lower than basic Levenshtein algorithm which achieves 100% upper accuracy level for single and two characters omission. However, basic Levenshtein achieves only 16% as a lower accuracy level.

The concepts used in this algorithm are general and can be applied to name matching in many other languages in which naming system is characterized by writing names in a restricted correct order, omission of one or more middle names, omission of non-significant components of names, and rare use of abbreviations for tokens. The only modification required to apply the proposed algorithm to other languages is the replacement of Arabic name frequency table with the specific language one.

As a future work, we are working to upgrade the algorithm to deal with cross-language name matching. Name matching across languages require approximate mapping of a name from one language into the other, and then applying a name matching algorithm. However, the problem is not a one to one character (or phoneme) replacement of name components in both languages, since different writing styles should be considered. For example, the current algorithm has to be modified to deal with name abbreviations, and change of token order in other languages.